\documentclass[conference]{IEEEtran}
\IEEEoverridecommandlockouts
\usepackage{cite}
\usepackage{amsmath,amssymb}
\usepackage{graphicx}
\usepackage{tikz}
\usepackage{pgfplots}
\usepackage{xcolor}
\usepackage{array}
\usepackage{booktabs}
\usepackage{url}
\usepackage[hidelinks]{hyperref}
\usetikzlibrary{arrows.meta,positioning,fit,calc}
\pgfplotsset{compat=1.18}
\def\BibTeX{{\rm B\kern-.05em{\sc i\kern-.025em b}\kern-.08em
    T\kern-.1667em\lower.7ex\hbox{E}\kern-.125emX}}
\newcommand{\smalltab}{\footnotesize\setlength{\tabcolsep}{2.8pt}\renewcommand{\arraystretch}{1.08}}

\begin{document}

\title{ContractHIL-HLS: Contract-Aligned Multi-Agent Workflow with Hardware-in-the-Loop Feedback for HLS Design}

\author{\IEEEauthorblockN{Jingbo Zhang$^{*}$, Haoxiang Sun$^{*}$, Wenbo Wang$^{*}$, Wenbo Zhang$^{\dagger}$}
\IEEEauthorblockA{Beijing University of Technology\\
Email: zhangjingbofpga@gmail.com, sunhaoxiang@emails.bjut.edu.cn, utzztu@emails.bjut.edu.cn, zhangwenbo@bjut.edu.cn\\
$^{*}$Equal contribution. $^{\dagger}$Corresponding author.}}

\maketitle

\begin{abstract}
This paper presents ContractHIL-HLS, a contract-aligned multi-agent workflow for practical HLS (high-level synthesis) engineering.
The workflow makes three contributions.
First, it introduces a structured contract as the semantic-alignment and task-execution artifact that translates natural-language requirements into explicit interfaces, constraints, validation checks, and rollback rules.
Second, it incorporates hardware information into the feedback loop by feeding HLS, Vivado, board runtime, power, and failure evidence back into generation, thereby extending LLM (large language model)-assisted HLS from kernel code toward system- and board-level closure.
Third, it decomposes agents by semantic lowering and execution tasks rather than by conversational roles: a Contract Agent lowers natural language into the contract, an HTML (Hypertext Markup Language) Agent renders it as structured HTML, and a Hardware-in-the-Loop Agent implements and revises the design with measured evidence.
We evaluate ContractHIL-HLS in two parts.
On 94 locally executable HLS-Eval tasks, the structured contract provides the largest small-design gain, improving the estimated single-sample testbench pass rate from 64.0\% to 70.2\%; the full flow reaches 70.4\% pass@1 and 76.6\% pass@5.
Because HLS-Eval does not exercise board-level design, we also validate ContractHIL-HLS on a board-tested PQC (post-quantum cryptography) secure-message accelerator, where the retained dual-bitstream organization reduces six-message average text runtime from 207.3~ms to 52.4~ms with positive routed timing slack on both images while preserving decrypted-message verification.
The implementation and evaluation artifacts are available at \href{https://www.github.com/BJUT-CS316-LAB/ContractHIL-HLS}{\nolinkurl{www.github.com/BJUT-CS316-LAB/ContractHIL-HLS}}.
\end{abstract}


\section{Introduction}
LLM-aided hardware design has matured through increasingly rigorous benchmarks, but practical HLS still requires workflows that preserve requirements, interfaces, EDA (electronic design automation) closure, host control, memory ownership, and FPGA (field-programmable gate array) deployment measurements across iterations. Chip-Chat studies interactive hardware design\cite{chipchat}, RTLLM evaluates RTL (register-transfer-level) generation\cite{rtllm}, and HLS-Eval extends evaluation to C/C++ HLS tasks\cite{hlseval}. These benchmarks show that models can generate hardware code, but they do not by themselves establish a workflow that keeps design intent aligned with tool feedback and deployment results.

ContractHIL-HLS formalizes such a workflow: it lowers natural language into a structured contract, retains that contract as inspectable HTML, and revises the implementation using hardware measurements. Agent surveys define agents through state, planning, action, workflow, and collaboration\cite{llmagentsurvey}\cite{multiagentsurvey}; we instantiate this perspective as an engineering loop rather than a role-label decomposition. The central question is whether an agent workflow can carry a design from intent to traceable implementation evidence rather than stop at prompt-local code generation.

HLS-Eval provides the quantitative small-design setting for that question, while PQC serves as the final hardware-in-the-loop engineering case because current PQC accelerators often emphasize operator or kernel innovation. Examples include KyberMat's NTT (number-theoretic transform)-oriented datapath\cite{kybermat} and CRYPHTOR's memory-unified CRYSTALS accelerator\cite{cryphtor}. We use PQC to test whether the same workflow can preserve system completeness, receiver-boundary rules, and rollback logic when the task expands from benchmark kernels to a board-tested secure-message system.

Our contributions are threefold.\\
\textbf{(1) Structured contract as an execution artifact:} we encode requirements as explicit interface, constraint, validation, optimization, and rollback fields so that generation is aligned to a shared engineering object rather than to loosely repeated prompt text.\\
\textbf{(2) Persistent HTML and hardware-in-the-loop feedback loop:} we render the contract into structured HTML and feed HLS, Vivado, and deployment measurements back into implementation and rollback decisions, turning tool output into first-class workflow state.\\
\textbf{(3) Agent coordination from benchmark tasks to an engineering case study:} we show that the same contract-aligned workflow improves HLS-Eval outcomes on 94 local tasks and then carries the design loop into a board-tested PQC secure-message system.

The rest of the paper is organized as follows. Section~II reviews background and related work, Section~III presents the workflow architecture and methodology, Section~IV reports HLS-Eval evidence, Section~V presents the PQC case study, and Section~VI concludes with limitations and future directions.

\begin{figure*}[!t]
\centering
\includegraphics[width=\textwidth]{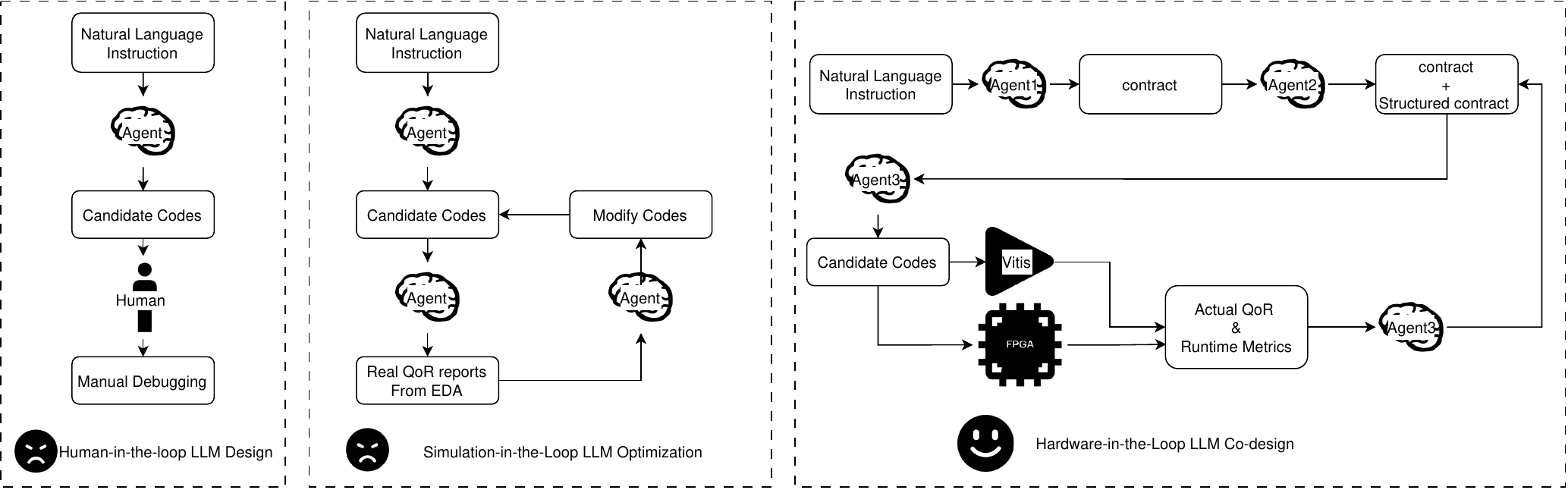}
\caption{ContractHIL-HLS workflow overview with retained contract state, EDA feedback, PYNQ (Python Productivity for Zynq) measurements, QoR (quality of results) metrics, and rollback decisions.}
\label{fig:workflow}
\end{figure*}

\section{Background and Related Work}
\subsection{LLM Agents and Artifact-State Decomposition}
Autonomous-agent work describes profile, memory, planning, and action\cite{llmagentsurvey}; multi-agent work emphasizes workflow and collaboration mechanisms\cite{multiagentsurvey}.
For hardware design, this separates \emph{what state is carried forward} from \emph{which conversational role a prompt emulates}.
Role prompts can support exploratory design, but an ``architect'' prompt may introduce constraints that a subsequent ``coder'' prompt omits.
ContractHIL-HLS instead assigns agents to typed transformations: natural language to contract, contract to HTML, and HTML plus EDA evidence to implementation or rollback.
The comparison target is therefore the direct-prompt or role-prompt flow whose carried-forward object is mainly conversation history.

\subsection{LLMs for RTL and Hardware Design}
Chip-Chat frames conversational co-design\cite{chipchat}, RTLLM and VerilogEval benchmark RTL generation\cite{rtllm}\cite{verilogeval}, VeriGen studies model specialization\cite{verigen}, and ChipGPT/ChipNeMo explore natural-language and domain-adapted chip tasks\cite{chipgpt}\cite{chipnemo}.
These works establish that LLMs can generate valid hardware artifacts, but their validation is usually code extraction, simulation, or module-level benchmark correctness.
ContractHIL-HLS builds on this evidence and shifts the question from isolated module generation to traceable workflow execution.

\subsection{LLMs for HLS and EDA Workflows}
HLS-oriented LLM work is more directly related to our setting because it considers C/C++ kernels, synthesis feedback, timing, QoR, and tests.
HLS-Eval provides benchmark tasks and an evaluation methodology for HLS code generation and editing\cite{hlseval}.
HLStrans explores C-to-HLS dataset construction\cite{hlstrans}, while SAGE-HLS uses syntax-aware AST (abstract syntax tree) guidance\cite{sagehls}.
These efforts are valuable but remain primarily kernel-centric so that parsing, compilation, testbench, and synthesis are reproducible.
They provide less coverage for bitstream partitioning, live-state ownership, PYNQ loopback, or cross-stage security boundaries.
Our evaluation therefore uses HLS-Eval for quantitative small-design evidence and uses PQC as a separate deployment stress test.

\subsection{PQC Hardware Acceleration}
\begin{figure}[!t]
\centering
\begin{tikzpicture}[x=0.51cm,y=0.48cm,>=Stealth]
\path (-1.08,-0.66) rectangle (11.50,6.78);
\draw[densely dotted,gray!32] (0,1.55) -- (8.7,1.55);
\draw[densely dotted,gray!32] (0,3.25) -- (8.7,3.25);
\draw[densely dotted,gray!32] (0,5.05) -- (8.7,5.05);
\draw[densely dotted,gray!32] (2.55,0) -- (2.55,6.25);
\draw[densely dotted,gray!32] (6.20,0) -- (6.20,6.25);
\draw[-{Stealth[length=3.7mm,width=2.4mm]},line width=0.95pt,draw=black] (0,0) -- (9.12,0);
\node[above,anchor=west,font=\bfseries\footnotesize] at (9.30,0.08) {Design scope};
\draw[-{Stealth[length=3.4mm,width=2.2mm]},line width=0.9pt,draw=black] (0,0) -- (0,6.6);
\node[above left,align=center,font=\bfseries\footnotesize] at (0,6.6) {Validation depth};
\foreach \x/\lab in {1.0/operator,4.4/kernel,7.9/system}
  \node[below,font=\footnotesize,align=center] at (\x,-0.04) {\lab};
\foreach \y/\lab in {0.9/code,2.4/synthesis,4.1/function,6.0/board}
  \node[left,font=\footnotesize,align=right] at (0,\y) {\lab};
\tikzstyle{pt}=[circle,draw=black!70,inner sep=1.7pt,line width=0.5pt]
\tikzstyle{lab}=[font=\footnotesize,align=center,text width=1.62cm,fill=white,inner sep=0.8pt]
\tikzstyle{callout}=[draw=gray!60,line width=0.42pt]
\node[pt,fill=blue!20] (ve) at (1.95,1.28) {};
\node[lab,anchor=west] (vel) at (2.27,1.12) {VerilogEval};
\draw[callout] (ve) -- (vel.west);
\node[pt,fill=blue!25] (rtl) at (3.35,2.05) {};
\node[lab,anchor=west] (rtll) at (3.78,2.18) {RTLLM};
\draw[callout] (rtl) -- (rtll.west);
\node[pt,fill=blue!35] (cc) at (4.85,2.82) {};
\node[lab,anchor=west] (ccl) at (5.36,2.94) {Chip-Chat};
\draw[callout] (cc) -- (ccl.west);
\node[pt,fill=green!25] (he) at (3.78,3.55) {};
\node[lab,anchor=east] (hel) at (3.34,3.98) {HLS-Eval};
\draw[callout] (he) -- (hel.east);
\node[pt,fill=green!35] (sh) at (5.78,4.53) {};
\node[lab,anchor=west] (shl) at (6.18,4.75) {SAGE-HLS};
\draw[callout] (sh) -- (shl.west);
\node[pt,fill=red!35,draw=red!65!black,inner sep=2.2pt,line width=0.8pt] (sc) at (8.05,5.72) {};
\node[lab,font=\footnotesize\bfseries,text width=2.55cm,anchor=east] (scl) at (7.62,6.28) {\mbox{ContractHIL-HLS}\\[-0.2ex]\textnormal{(ours)}};
\draw[callout] (sc) -- (scl.south);
\end{tikzpicture}
\caption{Conceptual positioning of representative LLM-assisted hardware workflows by design scope and validation depth.}
\label{fig:positioning}
\end{figure}
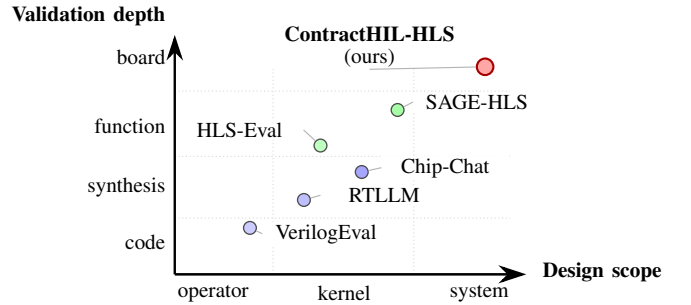

PQC is a demanding deployment case because ML-KEM (Module-Lattice-Based Key-Encapsulation Mechanism) and ML-DSA (Module-Lattice-Based Digital Signature Algorithm) combine compute-intensive transforms, memory-intensive packing, host orchestration, and security-sensitive dataflow\cite{fips203}\cite{fips204}.
Prior PQC work has produced strong fixed accelerators by specializing NTT, sampling, matrix-vector, and memory-unified datapaths\cite{kybermat}\cite{cryphtor}\cite{kid}\cite{pqcha}.
We use an ML-KEM/ML-DSA secure-message flow to test migration across interfaces, caches, bitstream choices, and validation boundaries, not to claim primitive-level latency superiority.
Fig.~\ref{fig:positioning} positions representative LLM-assisted hardware-code-generation work by design scope and validation depth.
ContractHIL-HLS targets the upper-right gap: system-level accelerator design filtered by synthesis, implementation, host control, and board measurements.
Across these lines of work, retained state is still usually prompt text or source-code snapshots, while HLS, Vivado, and deployment measurements rarely become first-class artifacts in the loop. This gap motivates the workflow architecture in Section~III.

\section{Workflow Architecture and Methodology}
\subsection{Workflow Overview and Agent Responsibilities}
ContractHIL-HLS is a three-agent workflow for transforming an imprecise user request into a traceable HLS implementation attempt.
The agents are distinguished by prompts and artifacts rather than by separate training.
The Contract Agent normalizes natural language into a structured engineering contract, the HTML Agent converts it into a structured HTML artifact, and the Hardware-in-the-Loop Agent uses the artifact plus hardware feedback to generate bounded HLS, Tcl (Tool Command Language), host-script, or documentation changes.
Independent scripts then parse, compile, perform CSim (C simulation), run CSynth (C synthesis), implement, measure, and score each candidate. Compatibility checks also reject prohibited use of the STL (C++ Standard Template Library).
Formally, the contract is
$C=\langle R,I,F,H,K,O,V,B\rangle$, where $R$ is role/platform, $I$ is the public interface, $F$ is functional intent, $H$ is hard and negative constraints, $K$ is compatibility with starter code and testbench, $O$ is optimization intent, $V$ is validation evidence, and $B$ is rollback policy.

\begin{table}[!t]
\caption{Prompt Contract Fields and Prevented Failures}
\label{tab:contractfields}
\centering
\smalltab
\begin{tabular}{p{0.25\linewidth}p{0.35\linewidth}p{0.24\linewidth}}
\toprule
Contract field & Prevented failure & Evidence/check\\
\midrule
Role/platform & unsupported flow or wrong tool assumptions & prompt metadata\\
Public interface & changed top function or signature & top-name/static check\\
Hard/negative constraints & dynamic allocation, STL, hidden loops & forbidden-pattern scan\\
Compatibility & missing headers and testbench mismatch & compile/testbench logs\\
Architecture reasoning & bitstream, phase, or data-boundary violations & design-rubric check\\
Measurement & accepting code without CSim/CSynth evidence & CSim/CSynth logs\\
Rollback & preserving changes that worsen metrics & score comparison\\
\bottomrule
\end{tabular}
\end{table}

\subsection{Contract-to-HTML Artifact Pipeline}
The Contract Agent does not write code.
It records role/platform, public interface, functional intent, hard constraints, compatibility requirements, optimization intent, validation checks, and rollback rules.
For HLS-Eval, these fields preserve the expected top function, starter code, and testbench-visible behavior; for board projects, they also record bitstream limits, phase boundaries, security rules, timing targets, and power metrics.
The HTML Agent then renders the contract as stable sections, tables, anchors, checklists, measurement fields, and rollback rules.
The Hardware-in-the-Loop Agent implements from that artifact and reports evidence.
Candidates are accepted only after local checks such as Can Parse, Can Compile, Can Pass TB (testbench), and Can Synth. Deployment candidates must also pass implementation and board measurements.

The workflow uses a concise prompt contract and a retained HTML contract.
HTML is dual-readable: engineers can inspect it, agents can consume its structured text, and scripts can parse or diff it.
The contract names rules such as \texttt{ap\_ctrl\_hs} control, DSA rejection-sampling boundaries, and decrypted-message verification.
For deployment checks, it records WNS (worst negative slack), LUT (lookup-table) and FF (flip-flop) utilization, DSP (digital signal processing) blocks, BRAM (block random-access memory), URAM (UltraRAM), dynamic power, board runtime, and rollback policy.
The loop consists of contract construction, HTML rendering, a bounded implementation request, the lowest-cost rejection check, an HLS/Vivado or benchmark check, and a keep-or-rollback decision from measured evidence.
For HLS-Eval, the loop stops at HLS-level evidence because the benchmark is kernel-level and board-neutral.
For deployment tasks, the same contract also requests Vivado block-design or IP (intellectual property) packaging notes and a PYNQ loopback script; in this paper, that board-loopback portion is empirically exercised only by the PQC case study.

\subsection{HIL (Hardware-in-the-Loop) Revision Loop and Platform Specialization}
The contract separates invariant rules, such as cryptographic dataflow and public interfaces, from platform choices such as bitstream splitting, cache placement, DSP binding, banking, and clock target.
This permits the same source project to prune caches on a small board or widen datapaths on a richer board without rewriting the evaluation logic.
Following prompt-pattern work on reusable specifications\cite{promptpatterns}, evaluation uses three regimes.
Direct is the original HLS-Eval-style task prompt; Contract adds the Contract Agent's structured fields; ContractHIL-HLS adds the HTML handoff and Hardware-in-the-Loop prompt.
For HLS-Eval, this deepest available loop ends at parse, compile, testbench, and HLS synthesis evidence; for PQC, it continues into Vivado implementation and PYNQ measurement.
Table~\ref{tab:contractfields} summarizes the contract fields and the failures they are intended to prevent.

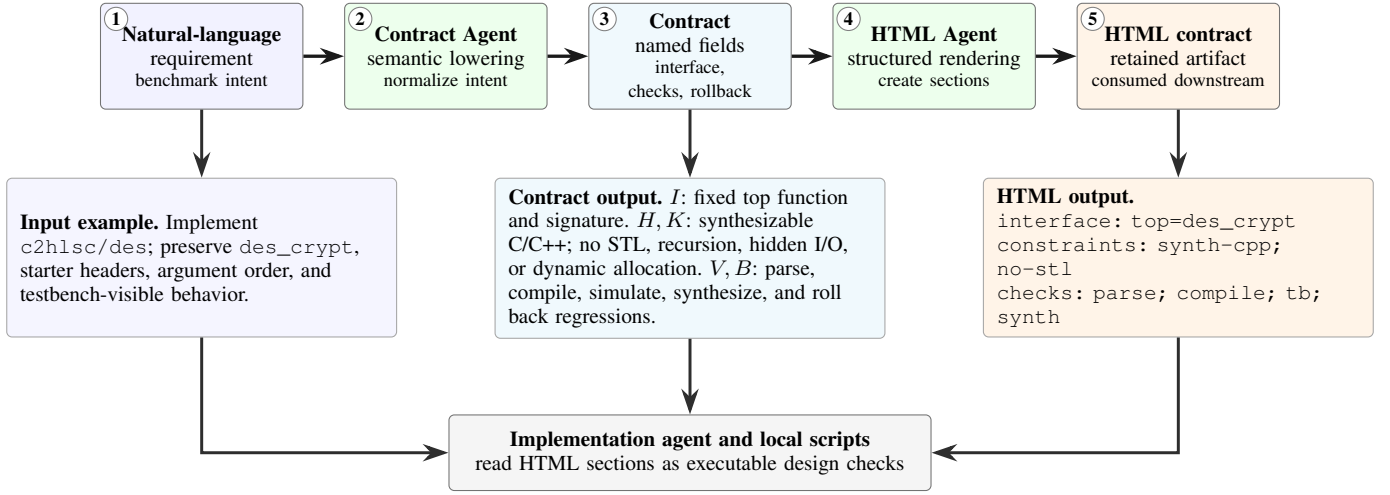
\begin{figure*}[!t]
\centering
\footnotesize
\begin{tikzpicture}[
  >=Stealth,
  stage/.style={draw=black!55,rounded corners=2pt,align=center,text width=0.136\textwidth,
    minimum height=14mm,inner xsep=3pt,inner ysep=3pt},
  data/.style={stage,fill=blue!5},
  agent/.style={stage,fill=green!7},
  contractbox/.style={stage,fill=cyan!5},
  html/.style={stage,fill=orange!9},
  card/.style={draw=black!35,rounded corners=2pt,align=left,text width=0.265\textwidth,
    minimum height=21mm,inner xsep=5pt,inner ysep=4pt,fill=#1},
  flowarrow/.style={-{Stealth[length=3.4mm,width=2.3mm]},line width=1.12pt,draw=black!82},
  mainarrow/.style={-{Stealth[length=3.2mm,width=2.2mm]},line width=1.25pt,draw=black!90},
  badge/.style={circle,draw=black!50,fill=white,inner sep=1pt,font=\scriptsize\bfseries}
]
\node[data] (nl) at (0,0) {\textbf{Natural-language}\\requirement\\[-1pt]\scriptsize benchmark intent};
\node[agent] (contractagent) at (3.23,0) {\textbf{Contract Agent}\\semantic lowering\\[-1pt]\scriptsize normalize intent};
\node[contractbox] (contract) at (6.46,0) {\textbf{Contract}\\named fields\\[-1pt]\scriptsize interface, checks, rollback};
\node[agent] (htmlagent) at (9.69,0) {\textbf{HTML Agent}\\structured rendering\\[-1pt]\scriptsize create sections};
\node[html] (htmlnode) at (12.92,0) {\textbf{HTML contract}\\retained artifact\\[-1pt]\scriptsize consumed downstream};
\foreach \n/\p in {1/nl,2/contractagent,3/contract,4/htmlagent,5/htmlnode}
  \node[badge,anchor=north west] at ($(\p.north west)+(1mm,-1mm)$) {\n};
\draw[mainarrow] (nl) -- (contractagent);
\draw[mainarrow] (contractagent) -- (contract);
\draw[mainarrow] (contract) -- (htmlagent);
\draw[mainarrow] (htmlagent) -- (htmlnode);

\node[card=blue!4,below=9mm of nl] (nlnote)
{\textbf{Input example.} Implement \texttt{c2hlsc/des}; preserve \texttt{des\_crypt}, starter headers, argument order, and testbench-visible behavior.};
\node[card=cyan!5,below=9mm of contract] (contractnote)
{\textbf{Contract output.} $I$: fixed top function and signature. $H,K$: synthesizable C/C++; no STL, recursion, hidden I/O, or dynamic allocation. $V,B$: parse, compile, simulate, synthesize, and roll back regressions.};
\node[card=orange!9,below=9mm of htmlnode] (htmlnote)
{\textbf{HTML output.}\par
\texttt{interface: top=des\_crypt}\par
\texttt{constraints: synth-cpp; no-stl}\par
\texttt{checks: parse; compile; tb; synth}};
\draw[flowarrow] (nl.south) -- (nlnote.north);
\draw[flowarrow] (contract.south) -- (contractnote.north);
\draw[flowarrow] (htmlnode.south) -- (htmlnote.north);

\node[stage,fill=gray!8,below=10mm of contractnote,text width=0.34\textwidth,minimum height=10mm] (consumer)
{\textbf{Implementation agent and local scripts}\\read HTML sections as executable design checks};
\draw[flowarrow] (nlnote.south) |- (consumer.west);
\draw[flowarrow] (contractnote.south) -- (consumer.north);
\draw[flowarrow] (htmlnote.south) |- (consumer.east);
\end{tikzpicture}
\caption{Natural-language-to-HTML flow in the contract-aligned workflow.}
\label{fig:stageexample}
\end{figure*}

Fig.~\ref{fig:stageexample} shows the semantic-lowering path used by the first two agents.
The Contract Agent does not attempt implementation; it translates the natural-language request into a bounded contract that explicitly names interface, constraint, validation, and rollback fields.
The HTML Agent then renders those fields into stable sections whose identifiers can be read by both the next agent and local scripts.
Thus the contribution is not simply a longer prompt, but the conversion of implicit natural-language requirements into stable, inspectable, and executable structure before code generation begins.

\section{Workflow Evaluation}
\subsection{HLS-Eval Setup}
All quantitative small-design evidence comes from the locally executable HLS-Eval subset, while the PQC design in Section~V is a separate deployment case.
We compare Direct, Contract, and ContractHIL-HLS with the same model endpoint and Vitis HLS toolchain, sampling each task five times.
Candidates are scored by Can Parse, Can Compile, Can Pass TB, and Can Synth; pass@1 is estimated from the average single-sample success rate across the five independent samples, and pass@5 reports whether any of five samples succeeds.
Can Synth is reported separately because a candidate may synthesize while failing the functional testbench.

\subsection{Main Results and Stage Ablation}
Table~\ref{tab:hlsevalstage1} shows that the dominant small-design gain comes from the structured contract: Contract raises Can Pass TB@1 from 64.0\% to 70.2\%.
ContractHIL-HLS reaches 70.4\% Can Pass TB@1 and 76.6\% Can Pass TB@5, but its single-sample synthesis estimate is lower.
Thus HLS-Eval is the contract-alignment test, while the PQC deployment is the HIL test for whether implementation and board measurements can alter a system-level design decision.

\begin{table}[!t]
\caption{HLS-Eval Full-Set Result and Published Context; pass@1 is the averaged single-sample estimate}
\label{tab:hlsevalstage1}
\centering
\scriptsize
\setlength{\tabcolsep}{1.8pt}
\renewcommand{\arraystretch}{1.05}
\begin{tabular}{p{0.31\linewidth}rrrrr}
\toprule
Method/model & Parse & Compile & TB@1 & TB@5 & Synth\\
\midrule
Direct & 97.9 & 82.1 & 64.0 & 73.4 & 94.9\\
Contract & 98.9 & 88.3 & 70.2 & 76.6 & 97.2\\
ContractHIL-HLS & 96.8 & 87.4 & 70.4 & 76.6 & 94.7\\
DeepSeek V3 zero-shot\cite{hlseval} & 100.0 & 94.1 & 63.3 & 65.9 & 93.2\\
\bottomrule
\end{tabular}
\end{table}

The first three rows are measured in our local harness.
The DeepSeek V3 row is included only to align metric definitions with the HLS-Eval paper's zero-shot context; it is not a controlled head-to-head workflow comparison.
For our rows, Can Synth includes independent \texttt{csynth\_design} runs for parseable candidates that failed the testbench.

\begin{figure}[!t]
\centering
\scriptsize
\begin{tikzpicture}
\begin{axis}[
    ybar,
    width=0.98\linewidth,
    height=44mm,
    ymin=0,ymax=90,
    ylabel={Rate (\%)},
    symbolic x coords={Direct TB1,Contract TB1,ContractHIL TB1,ContractHIL TB5,HLS-Eval TB5},
    xtick=data,
    xticklabel style={font=\scriptsize,rotate=18,anchor=east},
    nodes near coords,
    nodes near coords style={font=\scriptsize},
    bar width=7pt,
    enlarge x limits=0.18,
    ymajorgrids=true,
    xmajorgrids=false,
    tick align=inside,
    clip=true
]
\addplot[fill=blue!25] coordinates {(Direct TB1,64.0) (Contract TB1,70.2) (ContractHIL TB1,70.4) (ContractHIL TB5,76.6) (HLS-Eval TB5,65.9)};
\end{axis}
\end{tikzpicture}
\caption{HLS-Eval testbench-pass results. The final bar is the HLS-Eval paper's DeepSeek V3 zero-shot result under its original benchmark protocol.}
\label{fig:hlsevalbars}
\end{figure}
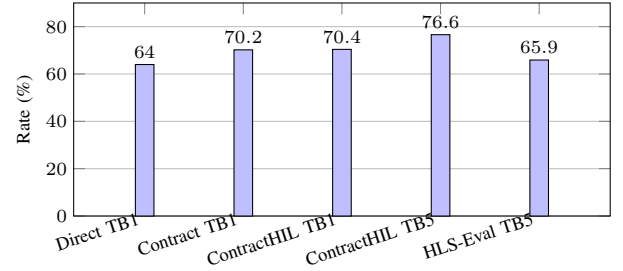

Fig.~\ref{fig:hlsevalbars} should be interpreted together with Tables~\ref{tab:hlsevalstage1} and~\ref{tab:nlcontract}.
The increase from Direct TB@1 to Contract TB@1 is the most direct quantitative evidence in the paper: the contract changes the estimated single-sample outcome before any board feedback is available.
The smaller movement from Contract TB@1 to ContractHIL-HLS TB@1 further constrains the claim, since HTML and the hardware loop do not automatically resolve kernel-level algorithm errors.
Accordingly, HLS-Eval supports the contract-alignment claim, while the PQC case evaluates hardware-in-the-loop system closure; the shared agent split connects the two experiments.
The token column in Table~\ref{tab:nlcontract} is also relevant for evaluating the workflow.
The Contract setting is slightly shorter than Direct because it replaces loosely repeated instructions with named fields, while ContractHIL-HLS spends more tokens to preserve the HTML handoff and implementation evidence.
Thus the small-design experiment establishes two implications: structured state can improve reliability without simply increasing prompt length, and the full hardware loop must be justified by system-level evidence rather than by HLS-Eval alone.

The same 94-case run also ablates the agents.
Direct tests the baseline prompt; Contract tests the Contract Agent alone; ContractHIL-HLS adds the HTML handoff and implementation-agent prompt.
This separates quantitative workflow evidence from the PQC deployment case study.

\begin{table}[!t]
\caption{ContractHIL-HLS Stage Ablation on HLS-Eval}
\label{tab:nlcontract}
\centering
\scriptsize
\setlength{\tabcolsep}{2.0pt}
\renewcommand{\arraystretch}{1.05}
\begin{tabular}{p{0.27\linewidth}ccrrr}
\toprule
Regime & Contract & HTML/loop & Pass TB@1 & Can Synth@1 & Tokens\\
\midrule
Direct & no & no & 64.0 & 94.9 & 3741\\
Contract & yes & no & 70.2 & 97.2 & 3555\\
ContractHIL-HLS & yes & yes & 70.4 & 94.7 & 4308\\
\bottomrule
\end{tabular}
\end{table}

\subsection{Coverage and Failure Analysis}
Table~\ref{tab:hlsevalcat} groups the same 94-task, five-sample run by benchmark family; Direct@1 is the Direct pass@1 estimate, and CHIL@1/CHIL@5 are the ContractHIL-HLS pass@1/pass@5 results.
Table~\ref{tab:hlsevalstage1} uses the underlying counts (301/470, 331/470, and 72/94), whereas Table~\ref{tab:hlsevalcat} rounds category rates to one decimal.
The full workflow improves the five-sample outcome for \texttt{c2hlsc} and improves the single-sample estimate on \texttt{chstone}, \texttt{pp4fpga}, \texttt{rosetta}, and \texttt{polybench}, but \texttt{machsuite} remains at 0.0\% even with five samples.
This negative result indicates that prompt structure improves reliability but cannot replace missing algorithmic capability.
For this reason, the paper does not present the contract as a universal code-generation solution.
It is a state-alignment mechanism: it preserves interface, constraints, and checks when the model is already close enough to produce a plausible implementation, and it exposes families where the model still lacks the required algorithmic transformation.

\begin{table}[!t]
\caption{HLS-Eval Category-Level Testbench-Pass Results from the Same 94-Task Run as Table~\ref{tab:hlsevalstage1}}
\label{tab:hlsevalcat}
\centering
\scriptsize
\setlength{\tabcolsep}{2.1pt}
\renewcommand{\arraystretch}{1.02}
\begin{tabular}{lrrrr}
\toprule
Cat. & n & Direct@1 & CHIL@1 & CHIL@5\\
\midrule
\texttt{c2hlsc} & 12 & 35.0 & 35.0 & 66.7\\
\texttt{chstone} & 20 & 80.0 & 95.0 & 100.0\\
\texttt{flowgnn} & 3 & 66.7 & 66.7 & 66.7\\
\texttt{gnnbuilder} & 3 & 100.0 & 100.0 & 100.0\\
\texttt{polybench} & 28 & 92.9 & 96.4 & 100.0\\
\texttt{pp4fpga} & 3 & 66.7 & 100.0 & 100.0\\
\texttt{rosetta} & 8 & 87.5 & 100.0 & 100.0\\
\texttt{machsuite} & 17 & 0.0 & 0.0 & 0.0\\
\bottomrule
\end{tabular}
\end{table}

Representative logs support the aggregate result.
In \texttt{c2hlsc/des}, the Direct candidate was parseable but failed integration, while the ContractHIL-HLS candidate passed C simulation and synthesis.
In \texttt{c2hlsc/block}, later pass@5 candidates passed after the first candidate missed a required \texttt{printf} declaration.
By contrast, \texttt{machsuite/aes\_aes} and the broader \texttt{machsuite} group stayed at 0/17, preventing an overbroad claim.
These coverage limits frame the final question of the paper: whether the same contract-aligned workflow still preserves correctness and rollback discipline when the task becomes a board-tested engineering case. Section~V addresses that question through the PQC case study.

\section{PQC Case Study}
This section uses PQC as a large engineering case study for the same contract-aligned workflow rather than as a second benchmark axis. ML-KEM and ML-DSA combine compute-intensive transforms, sender/receiver ownership boundaries, and board-level deployment constraints that are absent from HLS-Eval. These properties make PQC a useful demonstration of whether retained contract state and hardware-in-the-loop feedback can survive the transition from benchmark kernels to a board-tested secure-message system.

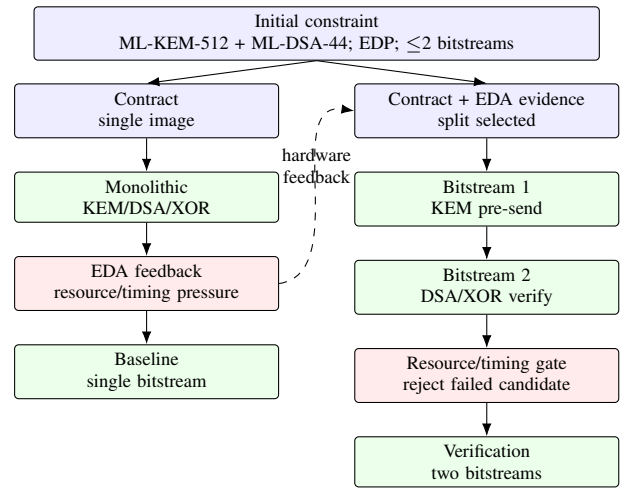
\begin{figure}[!b]
\centering
\scriptsize
\begin{tikzpicture}[node distance=4.4mm,>=Latex]
\tikzstyle{box}=[draw,rounded corners=1.5pt,align=center,text width=0.38\columnwidth,inner xsep=1.8pt,inner ysep=3.0pt]
\tikzstyle{keep}=[box,fill=green!8]
\tikzstyle{warn}=[box,fill=red!8]
\node[box,fill=blue!7,text width=0.83\columnwidth] (init) at (0,0)
{Initial constraint\\
ML-KEM-512 + ML-DSA-44; EDP; $\leq$2 bitstreams};
\node[box,fill=blue!7,below=3.0mm of init,xshift=-2.25cm] (sreq) {Contract\\
single image};
\node[keep,below=of sreq] (smono) {Monolithic\\
KEM/DSA/XOR};
\node[warn,below=of smono] (spressure) {EDA feedback\\
resource/timing pressure};
\node[keep,below=of spressure] (sout) {Baseline\\
single bitstream};

\node[box,fill=blue!7,below=3.0mm of init,xshift=2.25cm] (dreq) {Contract + EDA evidence\\
split selected};
\node[keep,below=of dreq] (dkem) {Bitstream 1\\
KEM pre-send};
\node[keep,below=of dkem] (ddsa) {Bitstream 2\\
DSA/XOR verify};
\node[warn,below=of ddsa] (dstop) {Resource/timing gate\\
reject failed candidate};
\node[keep,below=of dstop] (dout) {Verification\\
two bitstreams};

\draw[->] (sreq.south) -- (smono.north);
\draw[->] (smono.south) -- (spressure.north);
\draw[->] (spressure.south) -- (sout.north);
\draw[->,dashed] (spressure.east) to[out=8,in=172] node[above,font=\scriptsize,align=center]{hardware\\feedback} (dreq.west);
\draw[->] (init.south) -- (sreq.north);
\draw[->] (init.south) -- (dreq.north);
\draw[->] (dreq.south) -- (dkem.north);
\draw[->] (dkem.south) -- (ddsa.north);
\draw[->] (ddsa.south) -- (dstop.north);
\draw[->] (dstop.south) -- (dout.north);
\end{tikzpicture}
\caption{PQC contract loop under the initial constraint: implement ML-KEM-512 and ML-DSA-44, score by EDP (energy--delay product), and use at most two bitstreams. The first invocation builds a single-bitstream baseline; EDA feedback drives the second invocation to select a KEM/DSA-XOR split, verify two bitstreams, and reject later candidates that fail resource/timing gates.}
\label{fig:pqcloop}
\end{figure}

\begin{table*}[!t]
\caption{PQC case-study board summary on six text messages. Runtime excludes bitstream loading. KEM and DSA/XOR resource and power entries are per routed image; the dual-system power and EDP use the conservative sum of the two routed dynamic-power estimates.}
\label{tab:pqcsummary}
\centering
\scriptsize
\setlength{\tabcolsep}{2.2pt}
\renewcommand{\arraystretch}{0.98}
\begin{tabular}{@{}lccccc@{}}
\toprule
Metric & Processing-system software & Single baseline & Dual KEM image & Dual DSA/XOR image & Dual system\\
\midrule
Role & CPU (central processing unit) reference & monolithic PL (programmable-logic) bitstream & sender-side KEM & receiver-side DSA/XOR & 2-bitstream flow\\
Text avg. ms & 491.2 & 207.3 & -- & -- & 52.4\\
WNS (ns) & -- & 1.570 & 0.769 & 0.171 & both non-negative\\
LUT/FF & -- & 100857/77161 & 82862/37783 & 92051/38055 & per image\\
DSP/BRAM/URAM & -- & 218/144/36 & 720/72.5/35 & 287/100/10 & per image\\
Dynamic W & -- & 3.986 & 4.462 & 4.137 & 8.599\\
Text EDP mJ$\cdot$s & -- & 171.3 & -- & -- & 23.6\\
\bottomrule
\end{tabular}
\end{table*}

The reported averages are computed as
\[
\bar{t}_{\mathrm{text}}=\frac{1}{6}\sum_{i=1}^{6}\frac{1}{100}\sum_{r=1}^{100}t^{\mathrm{text}}_{i,r},
\]
where each \(t\) is one board-measured eight-operation secure-message run and bitstream loading is excluded.
With \(T\) in milliseconds, Table~\ref{tab:pqcsummary} reports \(\mathrm{EDP}=PT^2/1000\) in mJ$\cdot$s.
For the dual summary, \(\mathrm{EDP}_{\mathrm{dual}}=(P_{\mathrm{KEM}}+P_{\mathrm{DSA/XOR}})T_{\mathrm{dual}}^2/1000\), a conservative routed-report score that avoids inventing a composite power trace from two separately implemented images.
The dual resources in Table~\ref{tab:pqcsummary} are reported per bitstream rather than as a synthetic summed system utilization, because only one routed image is resident at a time.

The initial contract sets ML-KEM-512 and ML-DSA-44, EDP, and a two-bitstream limit. The first run produces a single-bitstream, end-to-end baseline and records its runtime, power, utilization, and timing state in the HTML contract. The second run selects the KEM/DSA-XOR split because it separates sender- and receiver-side live state. It is retained only when both images are correct, route successfully, and improve the conservative EDP score.

Table~\ref{tab:pqcsummary} gives the routed evidence. The monolithic image uses all 144 BRAM tiles despite 1.570~ns WNS. The split routes with 0.769~ns and 0.171~ns WNS, respectively, which keeps both images timing-legal while redistributing memory pressure.

Table~\ref{tab:pqcsummary} therefore evaluates the workflow rather than primitive latency. It retains the single-bitstream result as a measured baseline; the two-bitstream revision reduces text runtime from 207.3~ms to 52.4~ms and conservative EDP from 171.3 to 23.6~mJ$\cdot$s while retaining receiver-side validation. The HTML artifact records the selected organization, legal routed images, and rollback triggers. Together, HLS-Eval supplies reproducible small-design results, and PQC supplies a board-tested system case.

The contract also constrains the two system states. Before transmission, the KEM image produces sender-side data and writes values required after reload to host-visible storage. After reception, the DSA/XOR image reloads receiver-side inputs and validates only the decrypted message. This prevents a passing sender path from being mistaken for receiver validation and makes state ownership visible to implementation and review.

The benchmark and deployment evidence have deliberately different roles. HLS-Eval fixes C/C++ interfaces and testbenches over repeated samples, so it tests whether structured requirements improve local candidate reliability. It cannot test host orchestration, bitstream partitioning, or board-level timing. The PQC case tests these decisions with routed and runtime data but remains one deployment case. We therefore treat HLS-Eval as quantitative contract evidence and PQC as a system-level hardware-in-the-loop demonstration, not as a second benchmark or a primitive-latency claim.

The retained artifact narrows the revision surface. Each iteration records the interface and negative constraints, the validation gate, the report values, and the keep-or-rollback decision. A candidate that compiles but fails the testbench, timing, resource, or EDP gate cannot silently become the next context state. The resulting bitstream decision is inspectable because it follows a reported memory/timing condition and a receiver-boundary rule, rather than an unconstrained request for faster code.

Two reporting choices make the comparison conservative. Text runtime excludes bitstream loading for all reported message runs, whereas dynamic power is taken from routed reports for each image. The dual EDP therefore sums image powers even though the images are not resident simultaneously, and resources remain per image. The rollback rule rejects an organization that worsens its specified score or violates the functional boundary.
\vspace{-0.35\baselineskip}
\section{Conclusion}
\vspace{-0.15\baselineskip}
ContractHIL-HLS couples a structured contract and HTML handoff with a hardware-in-the-loop revision loop. On 94 HLS-Eval tasks, the contract raises testbench pass@1 from 64.0\% to 70.2\%; the full workflow reaches 70.4\% pass@1 and 76.6\% pass@5. The PQC case shows a board-tested, two-bitstream revision that reduces text runtime from 207.3~ms to 52.4~ms while retaining receiver-boundary validation. Future work will test more board-level designs and automate evidence checking.

\end{document}